\documentclass[12pt,a4paper,twoside]{article}
\makeatletter\if@twocolumn\PassOptionsToPackage{switch}{lineno}\else\fi\makeatother

\pdfoutput=1

\usepackage{pdfpages}

\usepackage{algpseudocode}
\usepackage{algorithm}
\usepackage{nccmath}
\usepackage{epsfig}
\usepackage{amsfonts,amssymb,amsbsy,latexsym,amsmath,tabulary,graphicx,times}
\usepackage[T1]{fontenc}
\usepackage{url,multirow,morefloats,floatflt,cancel,tfrupee,multicol}
\makeatletter
\usepackage{booktabs}
\usepackage{amssymb}

\AtBeginDocument{\@ifpackageloaded{textcomp}{\usepackage{textcomp}}}
\makeatother
\usepackage{colortbl}
\usepackage{pifont}
\usepackage[nointegrals]{wasysym}
\usepackage[superscript,biblabel]{cite}
\makeatletter

\@ifundefined{etal}{\def\etal{\textit{et~al}}}{}

\AtBeginDocument{
\expandafter\ifx\csname eqalign\endcsname\relax
\def\eqalign#1{\null\vcenter{\def\\{\cr}\openup\jot\m@th
  \ialign{\strut$\displaystyle{##}$\hfil&$\displaystyle{{}##}$\hfil
      \crcr#1\crcr}}\,}
\fi
}

\@ifundefined{tsGraphicsScaleX}{\gdef\tsGraphicsScaleX{1}}{}
\@ifundefined{tsGraphicsScaleY}{\gdef\tsGraphicsScaleY{.9}}{}
\def\checkGraphicsWidth{\ifdim\Gin@nat@width>\linewidth
	\tsGraphicsScaleX\linewidth\else\Gin@nat@width\fi}

\def\checkGraphicsHeight{\ifdim\Gin@nat@height>.9\textheight
	\tsGraphicsScaleY\textheight\else\Gin@nat@height\fi}

\def\fixFloatSize#1{}
\let\ts@includegraphics\includegraphics

\def\inlinegraphic[#1]#2{{\edef\@tempa{#1}\edef\baseline@shift{\ifx\@tempa\@empty0\else#1\fi}\edef\tempZ{\the\numexpr(\numexpr(\baseline@shift*\f@size/100))}\protect\raisebox{\tempZ pt}{\ts@includegraphics{#2}}}}

\AtBeginDocument{\def\includegraphics{\@ifnextchar[{\ts@includegraphics}{\ts@includegraphics[width=\checkGraphicsWidth,height=\checkGraphicsHeight,keepaspectratio]}}}

\DeclareMathAlphabet{\mathpzc}{OT1}{pzc}{m}{it}

\def\URL#1#2{\@ifundefined{href}{#2}{\href{#1}{#2}}}

\edef\fntEncoding{\f@encoding}

\makeatother

\newif\ifmultipleabstract\multipleabstractfalse%
\newenvironment{Figure}
  {\par\medskip\noindent\minipage{\linewidth}}
  {\endminipage\par\medskip}

\usepackage[hmargin=1.8cm,vmargin=2.2cm]{geometry}
\makeatletter
\def\author#1{\gdef\@author{\hskip-\dimexpr(\tabcolsep)\hskip1pt\parbox{\dimexpr\textwidth-1pt}{\centering #1}}}

\let\@articletype\@empty \def\articletype#1{\gdef\@articletype{{\fontsize{14}{16}\selectfont #1}}}

\usepackage{fancyhdr}

\fancypagestyle{headings}{\fancyhf{}\fancyhead[RE]{\MakeTextUppercase{\textbf{\RunningAuthor}}}\fancyhead[LE]{\textbf{\thepage}}\fancyhead[LO]{\MakeTextUppercase{\textbf{\RunningHead}}}\fancyhead[RO]{\textbf{\thepage}}}
\fancypagestyle{plain}{\fancyhf{}\fancyfoot[C]{\textbf{\thepage}}}

\AtBeginDocument{\usepackage{lastpage}}
\def\title#1{%
  \gdef\@title{%
    \ifx\@articletype\@empty\else\@articletype~\\\fi%
     #1}%
}

\usepackage{abstract}
\def\abstractname{\textbf{Abstract}}
\renewenvironment{onecolabstract}
{\vspace*{-.4pc}\trivlist\item[]\leftskip1pt\noindent\selectfont\hfill\abstractname\hfill\mbox{\null}\par\ignorespaces}{\endtrivlist}

\def\NormalBaseline{\def\baselinestretch{1.1}}

\usepackage{textcase}
\usepackage[explicit]{titlesec}
\titleformat{\section}[block]{\NormalBaseline\boldmath\bfseries}
{\thesection.}
{6pt}
{#1}
[]
\titleformat{\subsection}[hang]{\NormalBaseline\filright\itshape}
{\thesubsection.}
{6pt}
{#1}
[]
\titleformat{\subsubsection}[runin]{\NormalBaseline\filright\itshape}
{\hspace{16pt}\thesubsubsection}
{6pt}
{#1}
[]
\titleformat{\paragraph}[runin]{\NormalBaseline}
{\theparagraph}
{6pt}
{#1}
[]
\titleformat{\subparagraph}[runin]{\NormalBaseline}
{\thesubparagraph}
{6pt}
{#1}
[]

\titlespacing{\section}{0pt}{1.5\baselineskip}{.2\baselineskip}  
\titlespacing{\subsection}{0pt}{1.5\baselineskip}{.2\baselineskip}  
\titlespacing{\subsubsection}{0pt}{1.5\baselineskip}{.2\baselineskip}  
\titlespacing{\paragraph}{0pt}{.5\baselineskip}{10pt}  
\titlespacing{\subparagraph}{0pt}{.5\baselineskip}{10pt}

\usepackage{caption}
\DeclareCaptionLabelFormat{figlabel}{Figure #2}
\date{}
\makeatother

\usepackage{float}

\begin{document}

\title{Bacteria-inspired Robotic Propulsion from Bundling of Soft Helical Filaments at Low Reynolds Number}
\def\RunningHead{
Multi-flagella Robot
}
\def\RunningAuthor{Lim \etal}
\author{Sangmin Lim\thanks{Department of Mechanical and Aerospace Engineering, University of California, Los Angeles, California, USA.}, Achyuta Yadunandan \thanks{Department of Electrical and Computer Engineering, University of California, Los Angeles, California, USA.}, and M. Khalid Jawed\thanks{Department of Mechanical and Aerospace Engineering, University of California, Los Angeles, California, USA.}
}

\maketitle


{\begin{onecolabstract}
The bundling of flagella is known to create a "run" phase, where the bacteria moves in a nearly straight line rather than making changes in direction. Historically, mechanical explanations for the bundling phenomenon intrigued many researchers, and significant advances were made in physical models and experimental methods. Contributing to the field of research, we present a bacteria-inspired centimeter-scale soft robotic hardware platform and a computational framework for a physically plausible simulation model of the multi-flagellated robot under low Reynolds number ($\sim 10^{-1}$). The fluid-structure interaction simulation couples the Discrete Elastic Rods algorithm with the method of Regularized Stokeslet Segments. Contact between two flagella is handled by a penalty-based method. We present a comparison between our experimental and simulation results and verify that the simulation tool can capture the essential physics of this problem. Preliminary findings on robustness to buckling provided by the bundling phenomenon and the efficiency of a multi-flagellated soft robot are compared with the single-flagellated counterparts. Observations were made on the coupling between geometry and elasticity, which manifests itself in the propulsion of the robot by nonlinear dependency on the rotational speed of the flagella.

\def\keywordstitle{Keywords}
\end{onecolabstract}}
 
\begin{multicols}{2}

\section{Introduction}

Locomotion of micro-swimmers has drawn significant attention in biology and fluid dynamics since the 1950s~\cite{Gray1955,BergHowardC;Brown1972,Lighthill1975,Brennen1977,Purcell1977,Johnson1979}. On a microscopic scale, the physics of the fluid is dominated by viscosity over the inertial effect. A seminal paper by Purcell in 1977 explains that reciprocal motions do not provide propulsive force for microswimmers~\cite{Purcell1977}. Instead, several natural microswimmers use the polymorphic transformation of multiple or single slender filamentary appendages (e.g., cilia and flagella) to create nonreciprocal motion suitable for propulsion at low Reynolds number flow. Some cells, such as sperm and \textit{Vibrio cholerae}, utilizes single cilium and flagellum to move under a low Reynolds number. In contrast, a metachronal wave of ciliary arrays in humans and mammals and multiple flagella of \textit{Escherichia coli} exploits the interaction of multiple filamentary appendages with surrounding fluid for propulsion. Despite the differences in the number of appendages, many species of bacteria utilize the elastic helical flagellum/flagella as the main geometric structure to interact with the low Reynolds number flow. 
However, the mechanism of multi-flagellated locomotion and single-flagellated is fundamentally different. 

Multi-flagellated microswimmers have two modes of locomotion: ``run" and ``tumble"~\cite{BergHowardC;Brown1972,berg2003rotary,berg2008coli}. 
Run is a period of near straight line motion caused by the flagella's bundling. As multiple left-handed helix-shaped flagella turn in counterclockwise (CCW) direction for \textit{E. coli}, the flagella turn and synchronize to form a single or multiple bundles of helical shape~\cite{Turner2000}. On the other hand, a tumble is a period of random directional change caused by a change in the rotational direction of flagella, i.e., if single or multiple flagella of \textit{E. coli} rotates in clockwise (CW) direction. 
In essence, the multi-flagellated mechanism is an intricate interplay between geometric nonlinearity, hydrodynamics, and contact, contributing to robust bundling and direction-changing tumbling.

Inspired by the complexity of mechanics behind the simple driving mechanism, mechanical engineers also tried to formulate the motion of multi-flagellated bacteria~\cite{Kim2003,Kang2014,Ali2017,Danis2019}. Only recently, the bundling behavior was shown to be purely mechanical due to the interaction of the soft helical structures and the viscous fluid~\cite{Kim2003}. Further research on developing mechanical theory for flagellated locomotion beyond bundling and tumbling is also an active area of research. To fully understand the physics behind bacterial locomotion, the phenomenon such as synchronization and tangling of the bacterial flagella are being investigated~\cite{Tatulea-Codrean2020-sh,Tatulea-Codrean2021-nd,Tatulea-Codrean2022-fg}. Besides its complexity in physics, the multi-flagellated mechanism is vital from both robotic and biological perspectives due to the following features : (1) directional stability ~\cite{BergHowardC;Brown1972}, (2) redundancy of actuation~\cite{Mears2014},  (3) chemical secretion using flagella~\cite{Haiko2013,Ramos2004},  (4) improved efficiency in the swarm and propagation~\cite{Wolfe1989,Licata2016}

Compared to the multi-flagellated mechanism, locomotion used by bacteria with a single flagellum lacks interaction with one or more flagella, exploiting a similar yet different mechanism. Monotrichous bacteria such as \textit{Vibrio cholerae} exploit buckling instability induced by the hook of the flagellum to make a directional change in their motion~\cite{Son2013}.
Consequently, numerous biological findings~\cite{BergHowardC;Brown1972,Darnton2007,berg2008coli,berg2003rotary,SilvermanMicahel;Simon1977,Meadows2011,Minamino2011}, mechanical experiments~\cite{Rodenborn2013,Kim2003}, hydrodynamic theories for low Reynolds flow~\cite{Gray1955,Lighthill1975,Purcell1977,Purcell1997,Cortez2018}, and medical microbots~\cite{Edd2003,Beyrand2015,Ye2014,Zhang2009} explored and exploited such a mechanism. 

 Despite the differences in the mechanism of locomotion, both multi-flagellated and single-flagellated locomotion are related in that it is a crucial fluid-structure interaction prevalent in the microscopic world. 
 Prior works on soft robots actuated by flagella have considered simulation and experiments. To solve this fluid-structure interaction problem, the computational fluid dynamics model and slender body theory (SBT) were used to predict the motion of a single flagellated small-scale robot with rigid flagellum~\cite{Temel2014,Thawani2018}, ignoring the effect of flexibility of the flagella. With recent advancements in computational capability, the structural flexibility in a single-flagellated system can also be accounted for~\cite{Calisti2019,Huang2020,forghani2021control}; the flagellum can be modeled as a linear elastic Kirchoff rod~\cite{kirchhoff1859uber}. Multiple studies have demonstrated the modeling of multi-flagellated systems~\cite{Reichert2005,kim2004hydrodynamic,Reigh2012,Golestanian2011,Flores2005}. However, the coupling between long-range hydrodynamics, geometrically nonlinear deformation, and contact, has not been accounted for until recently\cite{huang2021numerical}.

In the field of microbots, several studies investigated the effect of multi-flagellated mechanisms. Due to the limited modes of locomotion that a single flagellated mechanism provide, Beyrand et al. presented multiple flagella microswimmers that can roll, run, and tumble ~\cite{Beyrand2015}. Ye et al. investigated multiple flagellated locomotion's benefits and the advantages of sinusoidal 2D geometry~\cite{Ye2014}. Even bio-hybrid microbots have been developed, created by assembling biological flagellated organisms with the artificial magnetic structure, showing remarkable results of controlled locomotion using magnetic field~\cite{Ali2017,Magdanz2013}. However, due to the nontrivial coupling between hydrodynamics, contact, and elasticity, researchers investigated lower-order coupling to solve the problem. e.g., (1) using rigid flagellum for a flagellated robot under viscous fluid ~\cite{Temel2014,Thawani2018}, (2) using a single elastic flagellum without modeling self-contact~\cite{Calisti2019,Huang2020} or (3) using ribbon-like multiple flagella without bundling behavior~\cite{Beyrand2015,Ye2014}. A comprehensive numerical model and physical prototype for flagella bundling still need further investigation. 

This paper presents a macroscopic soft robotic platform based on the propulsive mechanism of flagellated microorganisms and a physics-based computational framework to simulate the robot. The computational tool uses the Discrete Elastic Rods (DER) algorithm for elastic rod dynamics, Regularized Stokeslet Segments (RSS) method for hydrodynamics, including long-range interaction~\cite{Cortez2018}, and Spillman and Teschner's method of contact~\cite{Spillmann2008}. 
We first verify the simulation against the experiments with qualitative and quantitative comparisons. The simulation successfully captures the attraction between two flagella from hydrodynamic interaction.
Following a quantitative comparison between experiments and simulations, the model's shortcomings are discussed, and directions for future research are suggested. We find from experiments and simulations that the flagella buckling~\cite{Jawed2015} does not occur through purely hydrodynamic interactions between each flagellum for the multi-flagellated mechanism, while a single flagellated system may undergo buckling due to excessive hydrodynamic loads. Efficiency comparison between a single- and a multi-flagellated system shows that the single-flagellated robot has a slight efficiency advantage over its multi-flagellated counterpart. This simulation and the observations on the propulsive mechanism will set the foundation for further developing the soft robotic prototype. Furthermore, we report the experimental finding on cyclic unbundling without changing the velocity input. 

\section{Methods}
\label{sec:method}

\subsection{Experimental setup}

\label{experimentalsetup}
\begin{Figure}
    \centering
    \includegraphics[width=\columnwidth]{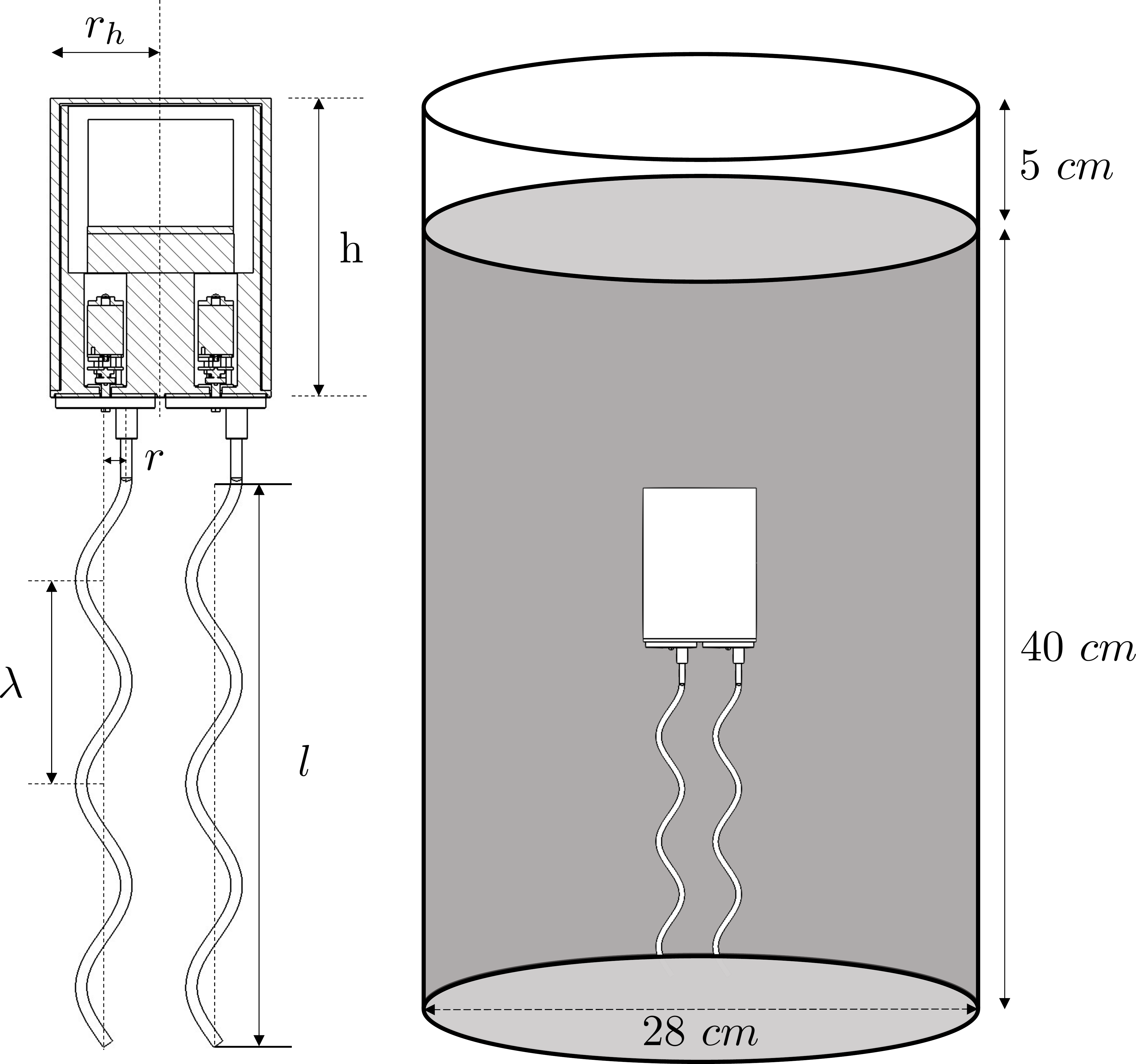}
    \captionof{figure}{Robot schematic, symbolic notations, and experimental setup. $r_h$ denotes the radius of the robot head, and $r$ denotes the radius of the helical flagellum. $\lambda$ denotes the helix pitch, $h$ denotes the height of the cylindrical head, and $l$ denotes the axial length of a flagellum. The robot tank was filled with glycerine in a cylindrical tank with a diameter of $28$cm and filled to $40$cm. }
    \label{setup}
\end{Figure}

For experimental data collection, we used glycerin as the viscous medium for our robot. A cylindrical tank with a diameter of $28$ cm and height of $45$ cm was used with glycerin filled up to a height of approximately $40$ cm. The robot was initially placed at the center of the glycerin tank to remain approximately $10$ cm apart from the sidewalls, as depicted in Fig.~\ref{setup}. For every experiment, temperature and viscosity were measured. Viscosity was measured using USS-DVT4 rotary viscometer, and the viscosity measurement was in good agreement with the nominal value of the glycerin; dynamic viscosity of $\mu = 0.956 \pm 0.2 $ Pa$\cdot$s. The temperature of glycerin was approximately $22^\circ$C throughout the experiment to minimize the effect of temperature on viscosity. The fluid was mixed thoroughly before the experiment to avoid variation in density inside the tank.

The robot head contained Wemos D1 mini microcontroller unit used for the motor control, two $3.7$ V $500$ mAh Lithium polymer (Lipo) batteries, and two mini geared DC motors. The motor was calibrated for angular velocity using Cybertech DT6236B Tachometer. Pulse width modulation (PWM) and rpm was calibrated within $\pm 1 (0.01\text{\%})$ rpm at $3.8$ V. The head is in cylindrical shape with radius of $3.1 \pm 0.01$ cm and height ($h = 8.2 \pm 0.01$ cm) and was built using fused deposition modeling (FDM) 3D printer with polylactic acid plastic (PLA) material ($\rho$ = $1.26$ g/cm$^3$). 
The robot head was comprised of the casing and the main body. Urethane wax was applied inside the casing, outside the main body, and in the motor chamber to further prevent the glycerin from penetrating the robot. The ballast was placed near the robot centroid for neutral buoyancy. The design of the robot was intentionally bottom-heavy to ensure stability. Flagella were attached to the bottom plates that are connected to the motors. 
\begin{Figure}
    \includegraphics[width=0.8\columnwidth]{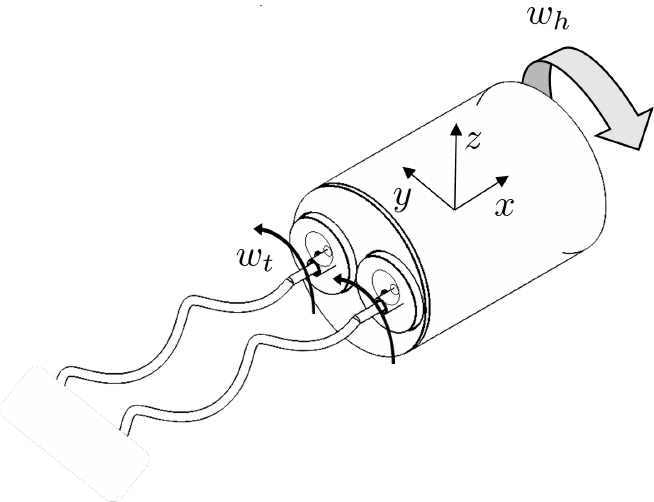}
    \captionof{figure}{Schematics of the robot. Rotational motions are exclusively denoted to show that both rotation of the head and rotation of the tail occurs. The body-fixed frame is shown in the center of the flagella robot to define the direction of rotation. The head rotates with angular velocity $\omega_h$ around the $x$-axis of the body-fixed frame, and tail rotates with angular velocity $\omega_t$ around the center of each helix.
    }
    \label{Forces}
\end{Figure}

For the flagella, Vinyl Polysiloxane (VPS) elastomeric material was used for fabrication with Young's modulus $E$ = $1255 \pm 49$ kPa and Poisson's ratio $\nu$ $\approx 0.5$ (i.e. nearly incompressible). Catalyst and base -- both liquid -- were mixed with a 1:1 volume ratio. Iron fillings were added to the liquid mixture to match the density of the glycerin ($\rho$ = $1.26$ g/cm$^3$). Left-handed helix shaped molds with different pitch parameters ($\lambda$ = $3.18, 4.45, 5.72$ cm) were 3D printed. Hollow Polyvinyl chloride (PVC) tubes were placed inside the molds, and the liquid mixture was injected inside the tubes. After waiting a few hours to cure, the PVC tubes were cut out, and filamentary soft helical rods were obtained~\cite{lazarus:hal-01447373}. More information on the choice of the geometry for both head and artificial flagella are available on \textbf{Supplementary methods section S1}.

The robot was activated from $30$ rpm to $70$ rpm at $10$ rpm intervals. The corresponding Reynolds number for the flagella ($Re = \frac{\rho ||\omega_T \times r|| r_0}{\mu}$), ranges from $0.0232$ - $0.0943$. the video of the robot submerged inside the glycerin tank for over $300$ seconds for each rotational speed was recorded for data collection. Out of the 300 seconds, we used the data between $30$ and $270$ seconds to ignore the initial transience during the speed ramp-up from 0 rpm to a prescribed total rpm, which is the sum of $\omega_t$ and $\omega_h$ depicted in Fig.~\ref{Forces}. Then, the videos were converted into jpg files with a frame rate of 1 frame per second. The image files were then processed using stacked image processing centroid calculation using the ImageJ image processing tool. 

\subsection{Physics-based simulation of the soft robot}
\label{simulation}

\begin{Figure}
    \centering
    \includegraphics[width=\columnwidth]{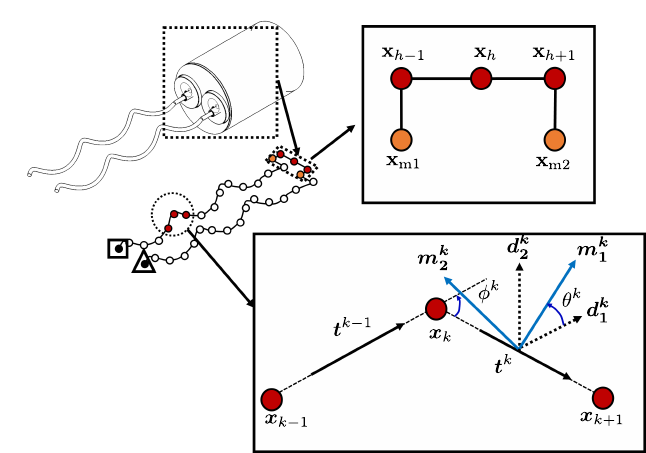}
    \captionof{figure}{Schematic showing discretization of the soft robot. The robot is modeled as a single rod starting from the left end of the flagella (denoted as a square) and ending at the right end of the flagella (denoted as a triangle). The head is modeled using three nodes (denoted as red dots). The center of the head is $\mathbf x_\textrm{h}$. The nodes to the left and right of the center are $\mathbf x_\textrm{h-1}$ and $\mathbf x_\textrm{h+1}$, respectively. Two orange dots, $\mathbf x_\textrm{m1}$ and $\mathbf x_\textrm{m2}$, are the nodes where the flagella are actuated.}
    \label{DER}
\end{Figure}

Fig.~\ref{DER} shows the discretized schematic of our robot. The rod is discretized into \textit{N} nodes:  $\mathbf{x}_k = \left[x_k , y_k , z_k\right]^T$ for $0 \leq k \leq N-1$. and \textit{N}-1 corresponding edges:  $\mathbf{e}^k = \mathbf{x}_{k+1} -\mathbf{x}_k$ for $0\leq{k}\leq {N-2}$. The degree of freedom (DOF) vector of discretized robot is defined as $\textbf{q} = \left[\mathbf{x}_0^T , \theta^0 , \mathbf{x}_1^T, \theta^1, ..., \mathbf{x}_{N-2}^T,\theta^{N-2},\mathbf{x}_{N-1}^T\right]^T$, 
where $\theta^k$ is the scalar twist angle at edge $\mathbf{e}^k$. Therefore, the size of the DOF vector for $N$ nodes is \textit{4N -1}. Hereafter, subscripts are used for the node-based quantities, and superscripts are used for the edge-based quantities.
\begin{table}[!htbp]
\centering
\resizebox{\columnwidth}{!}{%
\tabular{@{}l ccccc}
\\ 
    Parameter & Value(Exp/Sim) & Description \\    \hline
    r  & 0.0064 & Radius of helix (m)\\
    $\lambda$  & 0.0572  & Pitch of helix (m) \\
    l  & 0.0954  & Axial length of helix (m) \\
    $r_h$  & 0.031  & Radius of robot head (m)\\ 
    $r_0$  & 0.0016   & Radius of the rod (m)\\
    E  & 1.255 $\times 10^6$  & Young's Modulus (Pa)\\
    $\nu$  & 0.5  & Poisson's Ratio\\
    $\rho$ & 1260  & Density (kg/$\mathrm{m}^3$)\\ 
    $\mu$  & 0.956 $\pm$ 0.2 / 1.0  & Viscosity (Pa$\cdot$ s)\\ 
    $\triangle t$ & 1.0 $\times 10^{-4}$  & Time step  \\
    $\epsilon$& 1.67 $\times 10^{-4}$  & Regularization parameter\\    
    $|e|$ & 5.0 $\times 10^{-3}$ & Discretization length (m)\\ 
    $C_t$ & 4.8  & Translational drag coefficient\\ 
    $C_r$& 0.36  & Rotational drag coefficient\\ 
\endtabular%
}
\caption{Table of geometric, physical, and simulation parameters with symbol representations}
\label{parametertable}
\end{table}

An important characteristic of the DER method is the computation of the twisting of a rod simply by using a set of single scalar quantities $\theta^k$ embedded in the DOF vector. In this formulation, each edge has a reference frame (noted as $\mathbf{d_1^k}$, $\mathbf{d_2^k}$, $\mathbf{t^k}$ in Fig.~\ref{DER} that is orthonormal and adapted (i.e. $\mathbf{t^k}$ is the tangent along the $k$-th edge). The construction of the reference frame is first initialized at the first edge ($k=0$) at time $t=0$ with an arbitrary set of orthonormal vectors (with the condition that the third vector $\mathbf{t^0}$ is the tangent to the first edge). Then, the reference frame is parallel transported~\cite{Jawed2018} to the subsequent edges to form the reference frame on all the edges. After this initialization, the reference frame can be updated at each time step by parallel transporting $\mathbf{d_1^k}$, $\mathbf{d_2^k}$, $\mathbf{t^k}$ from the ``old" configuration (DOF vector before the time step) to the ``new" configuration (DOF vector after the time step). The material frame is also an adapted orthonormal frame (noted as $\mathbf{m_1^k}$, $\mathbf{m_2^k}$, $\mathbf{t^k}$ in Fig.~\ref{DER} that is identical to the reference frame at $t=0$. Since both the frames share a common director ($\mathbf{t^k}$), a single scalar quantity -- the twist angle, $\theta^k$ -- can be used to compute the material frame from the reference frame. An algorithmic representation of this update of the frame is shown in Algorithm 1. 

Based on the discretization, the elastic strains are required to calculate the energy and formulate the equations of motion (EOM) to march from time $t = t_i$ to time $t = t_{i+1} = t_i + \Delta t$, where $\Delta t$ is the time step size in the simulation outlined in Algorithm~\ref{alg:DER}. An elastic rod has three types of strains -- bending, twisting, and stretching -- associated with its deformation. We use the physical parameter presented in Table.~\ref{parametertable} for the calculation of the strains. Using these strains, we can calculate our system's stretching, bending, and twisting energy; the sum of three energy is noted as the elastic energy and is represented as Eq.~\ref{EqEnergy}. Details on the calculation of the strains and energy term can be found on \textbf{Supplementary methods section S2}.
\begin{equation}\label{EqEnergy}
    E_\textrm{elastic} = \underbrace{\sum_{k=0}^{N-2} E_k^s }_\textrm{stretching energy}+\underbrace{\sum_{k=1}^{N-2} E_k^b}_\textrm{bending energy}+\underbrace{\sum_{k=1}^{N-2} E_k^t}_\textrm{twisting energy},
\end{equation}

We can simply take the gradient of the energy terms with respect to the DOFs to get the elastic force at each DOF. The elastic force at the $k$-th DOF is $- \frac{\partial {E_\textrm{elastic}}}{\partial \mathbf{q}_k}$. The simulation marches forward in time by updating the configuration, i.e., DOF vector, of the robot based on EOM. We can even impart artificial configuration updates in the simulation that are dynamic. In particular, for actuation of the robot we implement a time-dependent fixed-rate natural twist on nodes $\mathbf x_\textrm{m1}$, and $\mathbf x_\textrm{m2}$ shown in Fig.~\ref{DER}. To propagate in time, the equation of motion to be solved at the $k$-th node is
\begin{equation}
    \label{EqMotion}
    \mathbf{f}_k \equiv \frac{m_k}{\triangle t}\left[\frac{\mathbf{q}_k(t_{i+1})-\mathbf{q}_k(t_i)}{\triangle t} - \mathbf{\Dot{q}}_k(t_i)\right]
    + \frac{\partial {E_\textrm{elastic}}}{\partial \mathbf{q}_k}  - \mathbf{f}_k^\textrm{ext} = 0,
\end{equation}
where $\mathbf q_k (t_i)$ is the old position (and the $k$-th element of the vector $\mathbf q (t_i)$), $\dot{\mathbf q_k}(t_i)$ is the old velocity, $m_k$ is the lumped mass at the $k$-th DOF, and $\mathbf{f}_k^\textrm{ext}$ is the external force on the $k$-th DOF. Note that Eq.~\ref{EqMotion} is simply a statement of Newton\rq{}s second law. External forces may include gravity, contact, and hydrodynamics, and the $(4N-1)$-sized external force vector can be written as 
\begin{equation}
    \mathbf{f}^\textrm{ext} = \mathbf {f}^\textrm{h} + \mathbf {f}^\textrm{head} + \mathbf {f}^\textrm{c},
    \label{eq:extForce}
\end{equation}
where $\mathbf {f}^\textrm{h}$ is the hydrodynamic force vector on the flagella, $\mathbf {f}^\textrm{head}$ is the hydrodynamic force vector on the head, and $\mathbf {f}^\textrm{c}$ is the contact force vector.  (\textbf{Supplementary methods section S3})

Now that the EOM is defined, we need to solve the system of $(4N-1)$ equations defined by Eq.~\ref{EqMotion} to compute the new position vector, $\mathbf{q} (t_{i+1})$. The Newton-Raphson method can be used to solve the equations which require the Jacobian of Eq.~\ref{EqMotion}. The $(k,m)$-th element of the square Jacobian matrix is   
\begin{equation}
\label{Jac:eq}
    \mathbb{J}_{km} = \frac{\partial \mathbf{f}_k}{\partial \mathbf{q}_m} = \mathbb{J}_{km}^\mathrm{inertia}+\mathbb{J}_{km}^\mathrm{elastic}+\mathbb{J}_{km}^\mathrm{ext} ,
\end{equation}

    \begin{algorithm}[H]
    \caption{Multiflagella soft robot simulation}
    \label{alg:DER}
    \begin{algorithmic}[1]
        \Require $\boldsymbol{q}(t_j),\Dot{\boldsymbol{q}}(t_j)$ 
        \Require $(\boldsymbol{d_1^k}(t_j),\boldsymbol{d_2^k}(t_j),\boldsymbol{t^k}(t_j))$
        \Ensure $\boldsymbol{q}(t_{j+1}),\Dot{\boldsymbol{q}}(t_{j+1})$
        \Ensure $(\boldsymbol{d_1^k}(t_{j+1}),\boldsymbol{d_2^k}(t_{j+1}),\boldsymbol{t^k}(t_{j+1}))$
        \State {Guess : $\boldsymbol{q}^{(1)}(t_{j+1})\xleftarrow{}\boldsymbol{q}(t_j)$}
        \State {$n\xleftarrow{}1$}
        \State {Calculate $\mathbf{f}^h$ and $\mathbf{f}^\mathrm{head}$} \\
        \Comment{\textbf{Supp. methods - sec. S3}}
        \State{solved$~\xleftarrow{}0$}
        \While{solved == 0}
        \While{error > tolerance}
        \State {Compute ref. frame} using {$\boldsymbol{q}^{(n)}(t_{j+1})$} 
        \State  {$(\boldsymbol{d_1^k}(t_{j+1}),\boldsymbol{d_2^k}(t_{j+1}),\boldsymbol{t^k}(t_{j+1}))^{(n)}$}
        \State {Compute ref. twist} {$\triangle m_\mathrm{k,ref}^{(n)}$}
        \State {Compute material frame}
        \State  {$(\boldsymbol{m_1^k}(t_{j+1}),\boldsymbol{m_2^k}(t_{j+1}),\boldsymbol{t^k}(t_{j+1}))^{(n)}$}
        \State {Compute \textbf{f} and $\mathbb{J}$} \Comment{Eq.~\ref{EqMotion},~\ref{Jac:eq}}
        \State {{$\triangle \boldsymbol{q}$} {$\xleftarrow{}$} $\mathbb{J}${$\backslash$} {$\textbf{f}$}}
        \State {$\boldsymbol{q}^{(n+1)}$}  $\xleftarrow{}$ {$\boldsymbol{q}^{(n)}$}  - {$\triangle \boldsymbol{q}$} 
        \State{error {$\xleftarrow{}$} sum ( abs ({$\textbf{f}$} $) )$ }
        \State{$n$ {$\xleftarrow{}$} $n+1$}
        \EndWhile
        \State{solved$\xleftarrow{}$1}
        \For{$l=0$ to $l=N-2$}
        \For{$m=0$ to $m=N-2$}
        \State{Compute $\delta^\mathrm{min}_{(l,m)}$}
        \If{$\delta^\mathrm{min}_{(l,m)}< 2r_0$}
        \State{Compute $\mathbf{f}^c_l,\mathbf{f}^c_{l+1},\mathbf{f}^c_{m},\mathbf{f}^c_{m+1}$}
        \State{} \Comment{\textbf{Supp. methods - sec. S4 }}
        \State{solved$~\xleftarrow{}$0}
        \EndIf
        \EndFor
        \EndFor
        \EndWhile
        \\
    {$\boldsymbol{q}(t_{j+1})\xleftarrow{}\boldsymbol{q}^{(n)}(t_{j+1})$}\\
    {$\Dot{\boldsymbol{q}}(t_{j+1})\xleftarrow{}\frac{\boldsymbol{q}(t_{j+1})-\boldsymbol{q}(t_{j})}{\triangle t}$}\\
    {$(\boldsymbol{d_1^k}(t_{j+1}),\boldsymbol{d_2^k}(t_{j+1}),\boldsymbol{t^k}(t_{j+1}))\xleftarrow{}(\boldsymbol{d_1^k}(t_{j+1}),\boldsymbol{d_2^k}(t_{j+1}),\boldsymbol{t^k}(t_{j+1}))^{(n)}$}
    \end{algorithmic}
    \end{algorithm}

The expressions for the Jacobian terms associated with the elastic forces are available in the literature~\cite{Jawed2018}. The Jacobian terms associated with some external forces ($\mathbf {f}^\textrm{h}, \mathbf {f}^\textrm{head}$) cannot be analytically evaluated and those terms are simply set to zero. In other words, those forces are incorporated into the simulation in an Euler-forward fashion.

After solving Eq.~\ref{EqMotion} to calculate the new position $\mathbf{q}_k(t_{i+1})$, new velocity  can be trivially computed from $\dot{\mathbf{q}}_k(t_{i+1}) = \left( \mathbf{q}_k(t_{i+1}) - \mathbf{q}_k(t_{i}) \right) / \Delta t$.

\section{Results and Discussion}
\subsection{Validation of Physics-based Simulation of Multi-flagellated robot}\label{sec:validation}
\begin{figure*}[!htbp]
    \centering
    \includegraphics[width=\textwidth]{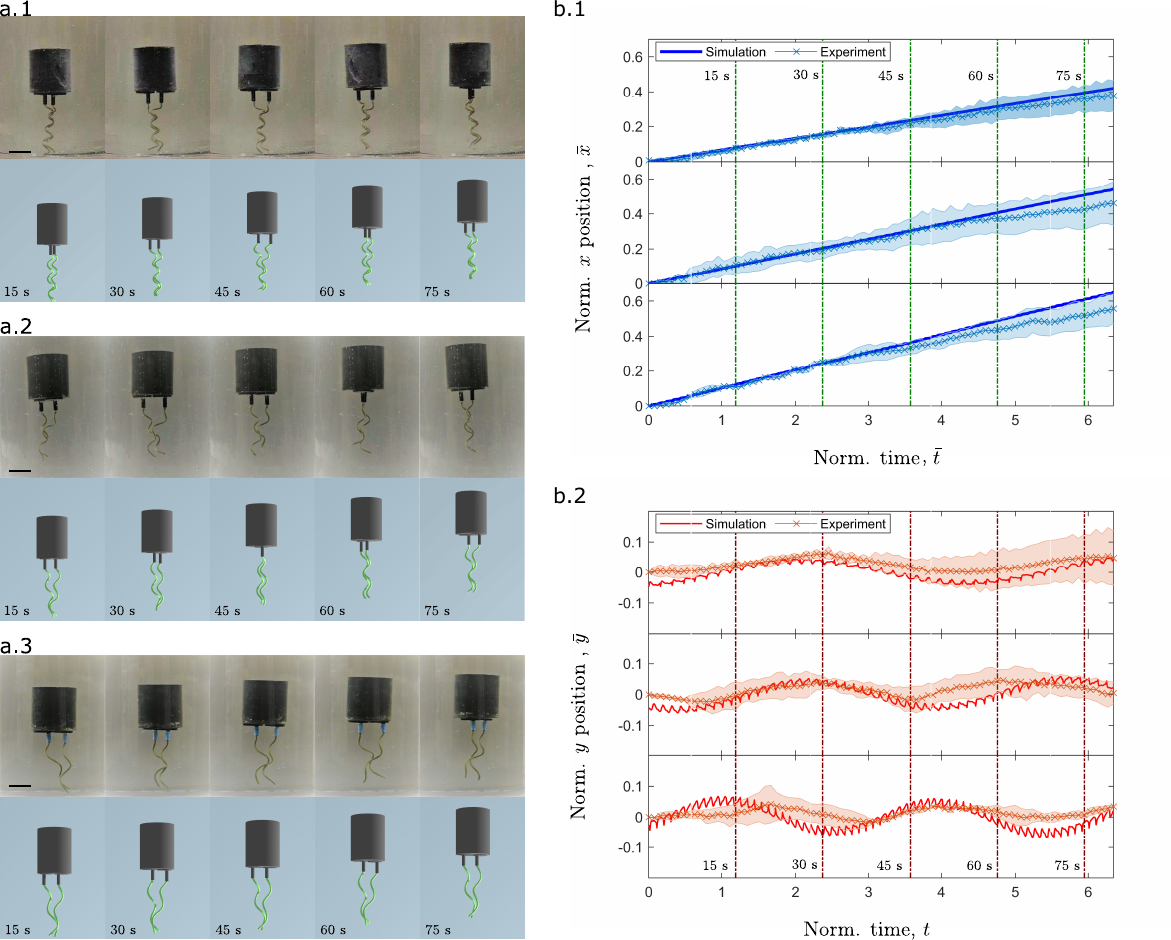}
    \captionof{figure}{Comparison of data for experiment and simulation for different pitch to helix radius ratio of flagella ($\bar{\lambda} = 5, 7, 9$) : (a) Snapshot comparison between experiment and simulation at $60$ rpm, scale bar: $3$cm. The deformation of elastic filamentary flagella in experiment and simulation has a plausible agreement ; (b) Comparison of experimental data with simulation at $30, 40, 50$ rpm for $\bar{\lambda} = 7$. (b.1) compares normalized x position, and (b.2) compares normalized y position against normalized time. The shaded area represents the standard error of experiment data. Normalized time for the snapshot in Fig 5. (a) is drawn as dotted lines with time notations in Fig 5. (b).}
    \label{positiondata}
\end{figure*}
To demonstrate our simulation model's validity, we compared our multi-flagellated robot's locomotion against the simulation. For generality, we present our results in a nondimensional form. Due to the slender geometry of the system, bending is the dominant deformation mode. Balancing the elastic bending force and the external viscous loading yields a characteristic time scale of $\mu l^4 / (EI)$~\cite{Jawed2015,Jawed2016}. This characteristic time is used to nondimensionalize the time, and a characteristic bending force of $EI/l^2$ is used to nondimensionalize the forces. The distance was nondimensionalized with the axial length. Hereafter, overbar ($\bar{\;}$) represents normalized quantities, i.e. $\bar{t} =  \frac{t EI}{\mu l^4}$, $\bar{\omega} =  \frac{\omega \mu l^4}{EI}$ , $\bar{v} = \frac{v \mu l^3 }{EI}$ , $\bar{F_p} = \frac{F_p l^2 }{EI}$, $\bar{x} = \frac{x}{l}$,
$\bar{\lambda}=\frac{\lambda}{r}$, etc. 

Fig.~\ref{positiondata}.a shows snapshots from experiments and simulations at three different values of pitch: $\bar \lambda = \{ 5, 7, 9\}$. From Fig.~\ref{positiondata}.a, we first qualitatively analyze the match between the transitional motion of the flagella crossing and bundling behavior. We noted during experimental observations that the $\bar{\lambda} = 5$ case formed a bundle throughout the entire range of angular velocity variation ($30$ to $70$ rpm). On the other hand, $\bar{\lambda} = 7$ case formed a partial bundle at $50$ and $60$ rpm, and $\bar{\lambda} = 9$ did not bundle but had continuous contact between two flagella. In Fig.~\ref{positiondata}.b, we present quantitative comparison of experimental data with simulation at $30, 40,$ and $50$ rpm for $\bar{\lambda} = 7$ and plot the position of the robot along $x$ and $y$ directions against normalized time. The data frequency of the experiment is one frame per second (fps), and the data frequency of the simulation is ten fps. 

We obtain the values of the numerical prefactors $C_t$ and $C_r$ by data fitting using data obtained on experiment with $\overline{\lambda} = 7$.  Details on the numerical prefactors are found on \textbf{Supplementary methods section
S3}. By comparing the mean total least squared error for the axial velocity v. rotational velocity, the prefactors that provided minimal error between the experiment and simulation were used. Both prefactors are the coefficients to account for the non-spherical shape of the robot body in translational, rotational hydrodynamic drag calculation based on the sphere at Stokes flow. 

The experiment and simulation results show reasonable agreement in the positional data for $\bar{x}$ and $\bar{y}$ in Fig.~\ref{positiondata}.b. The normalized $x$ position, $\bar{x}$, shows a better match between the simulation and experiment (Fig.~\ref{positiondata}.b.1). The normalized $y$ position, although the experiment follows similar oscillatory trend, the experimental robot exhibited smaller average magnitude with higher standard deviation. From Fig.~\ref{positiondata}.b.2, we can observe that as the angular velocity increases, the oscillation frequency in normalized $y$ position increases accordingly. This represents an undulatory sideways motion of the robot as it moves upward (along $x$ direction). Even though both experiments and simulations show the same trend, there exist discrepancies in the higher rpm and lower normalized pitch cases. We attribute this to the friction between the flagella surfaces. Unlike simulation, where we resolve contact between the two flagella without consideration of friction between each flagellum, we observed in our experiment that once it partially or fully bundles, the bundled part has a high frictional force that makes the flagella be kept in the bundled configuration when transitioning to the partially bundled regime.
\begin{Figure}
    \centering
    \includegraphics[width=\columnwidth]{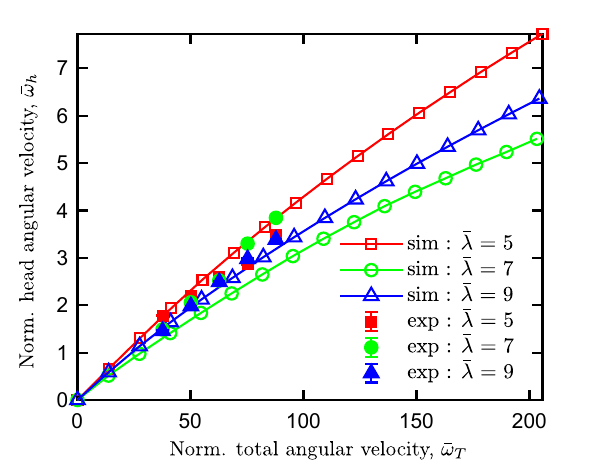}
    \captionof{figure}{Comparison of simulation and experiment for head and total angular velocity. The experimental data are in symbols with error
bars.}
    \label{head_exp}
\end{Figure}
Simple moment balance tells us that the cell body (i.e., the head of the robot) and the flagella have to rotate in opposite directions for net zero torque. It is known through previous works that the counter-rotation of the cell body in bacterial locomotion contributes to the trajectory and efficiency of the organism and could even contribute to the bundling of the flagella.~\cite{Powers2002,Constantino2016} We use our robot to investigate the rotation of the head and, in Fig.~\ref{head_exp}, plot the angular velocity of the head as a function of the total angular velocity from both experiments and simulations at three different values of the pitch. The error bar is obtained from the standard deviation of the experimental values. At lower values of total angular velocity ($\bar \omega_T \lesssim 100$), there is little variation in the angular velocity of the head at different pitch values. In both experiments and simulations, we see that the angular velocity of the head increases almost linearly with the total angular velocity. We exploit our simulations to probe the higher angular velocity regime and clearly see that the head angular velocity increases sublinearly with the total angular velocity. The variation in the head angular velocity as a function of the pitch of the flagella is worth mentioning. Among the three examined here ($\bar \lambda = \{ 5,7,9\}$), the head angular velocity is the highest at $\bar \lambda = 7$. This nonlinear dependence on the flagella's geometric parameter (pitch) may be counterintuitive; however, it is a manifestation of the problem's highly nonlinear and coupled nature. This type of nonlinearity with a variation of geometry for a rigid helical structure can also be found in ~ Ref. \cite{Rodenborn2013}, which used a single-flagellated system and analyzed the normalized force with respect to the normalized pitch. However, from our experiment and simulation, we observed that the coupling of the elasticity results in a different nonlinear pattern in the velocity of the robot, which is relevant to the forces in the direction of propulsion from the previous work done with a rigid structure. This result shows a new outlook for future research to determine the relationship between the elasticity, bundling, and hydrodynamic effect that changes previously known geometric dependency on the force and torque for a single rigid helical structure.

\begin{Figure}
    \centering
    \includegraphics[width=\columnwidth]{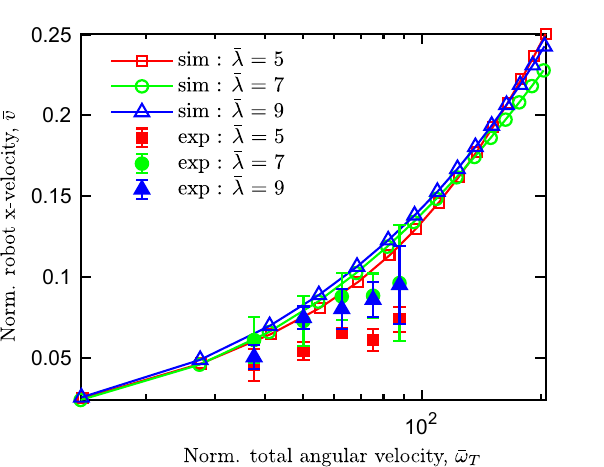}
    \captionof{figure}{Nonlinear relationship of angular velocity and robot x-velocity captured through simulation. The experimental data are in symbols with error bars.}
    \label{vel_exp}
\end{Figure}

Next, in Fig.~\ref{vel_exp}, the velocity of the bacteria robot along the $x$-axis is shown as a function of the normalized total angular velocity, $\bar{\omega}_T$. The error bar represents the starndard deviation of the normalized $x$-velocity. The $x$-velocities are obtained through the coefficient of the linear fitting of the position value through MATLAB polyfit function. Interestingly, the translational velocity increases superlinearly with the total angular velocity in the regime explored here. The nonlinear dependence of translational velocity on the pitch of the flagella is also apparent. The case with $\lambda=7$ results in lower propulsive speed than the $\lambda=5$ and $\lambda=9$ cases. In Fig.~\ref{vel_exp}, the simulation results overestimate the velocity of the experimental robot. %

This discrepancy in simulation and experiment can be attributed to the friction between the two flagella that was observed in experiments. Since the amount of contact is more dominant at lower pitch values, the experiments and the simulations differ further for the case with $\bar{\lambda}= 5$ compared with $\bar \lambda=7$ and $\bar \lambda=9$ cases. At this stage of our research, the simulation tool enforces non-penetration conditions but does not incorporate friction. This implies that one flagellum can smoothly slide past another flagellum without any resistance from friction. However, that is not the case in the real world. In experiments, the flagella form a tighter bundle compared with simulations, leading to a lower net propulsive force forward. Incorporating physically-accurate friction inside a low Reynolds environment is a direction of future research. The simulation tool presented in this paper, which models the entire system as a single rod for computational efficiency, can be used to explore various models of friction and eventually formulate an accurate model that matches experiments. A few recent works have explored friction among rods in simpler settings~\cite{li2020incremental, choi2021implicit}, and our simulations can be augmented to include such friction models.

\subsection{Comparison with Single-Flagellated Robot}
\label{sec:comparison_with_single_flagellum_robot}


In this section, we take one step toward a mechanistic understanding of the difference between these two modes of locomotion -- single-flagellated and multi-flagellated. Locomotion of a robot (or bacterium) with a single flagellum was recently investigated using a DER-based numerical framework~\cite{Huang2020}. A single-flagellated robot cannot exhibit bundling; however, it can undergo buckling instability beyond a critical value of the total angular velocity when the resulting external hydrodynamic force is too large. In Fig.~\ref{singlecompare}, we utilize our same simulation tool to model a single-flagellated robot. 
\begin{Figure}
    \centering
    \includegraphics[width=\columnwidth]{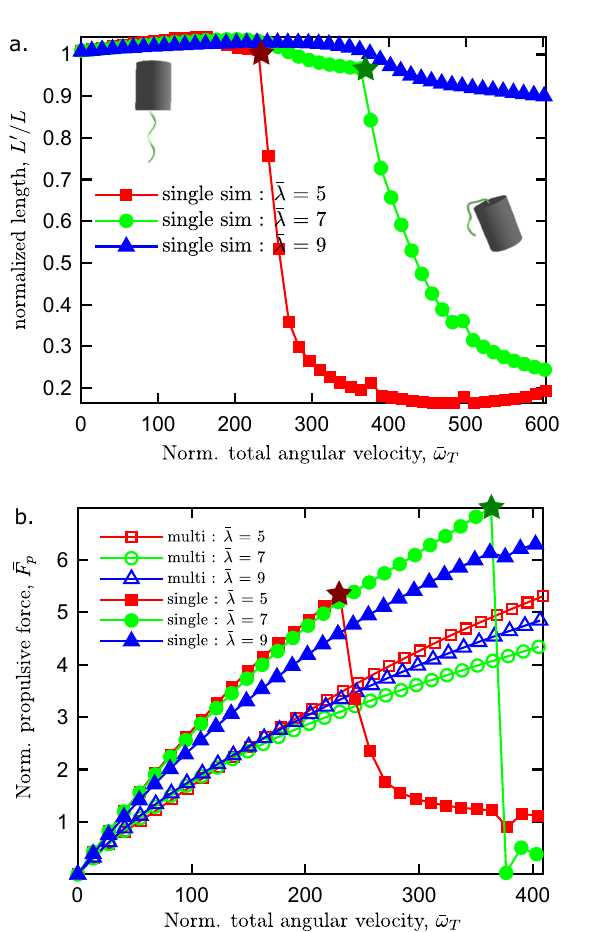}
    \captionof{figure}{comparison between single-flagellated robot and multi-flagellated robot simulation: (a) Plot of a normalized tail to head distance with respect to the normalized angular velocity. Star symbols represent the critical angular velocity where the flagellum buckles. Rendered image of unbuckled (left) and buckled (right) state of the robot shown within the graph. (b) A figure of normalized propulsive force with respect to the normalized angular velocity. Star symbols represent the buckling.}
\label{singlecompare}
\end{Figure}

We assumed that all the parameters were the same between the single-flagellated robot and the multi-flagellated case discussed above. The only difference is the number of flagella. Fig.~\ref{singlecompare}.a shows the Euclidean distance, $L\rq{}$, between the head node and the tail node (the node at the free tip of the flagellum) as a function of the total angular velocity, $\omega_T$, obtained from simulations. This apparent length, $L\rq{}$, has been normalized by its value at $\omega_T=0$ so that all the curves for three different pitch values start at $(0,1)$. The star symbols represent the angular velocity beyond which the apparent length, $L\rq{}$, of the robot abruptly drops, and the flagellum buckles. Two snapshots -- one of an unbuckled configuration and one of a buckled shape -- are also shown in Fig.~\ref{singlecompare}.a. For $\bar{\lambda} = 5$, the flagella buckles at $\bar{\omega}_T \approx 230$ and the $\bar{\lambda} = 7$ case buckles at $\bar{\omega}_T \approx 363$. The case for $\bar{\lambda} = 9$ does not buckle in the regime explored in this figure. The findings on the single-flagellated robot are similar to the study in Huang et al.~\cite{Huang2020}.

\begin{Figure}
    \centering
    \includegraphics[width=\columnwidth]{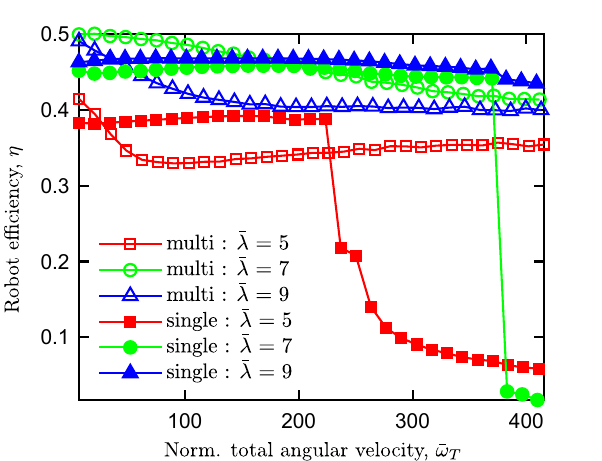}
    \captionof{figure}{Efficiency graph for the single and multi flagella robot simulation. Efficiency is defined as the ratio of the force and torque of the head. Due to flagella interaction, the multi-flagellated robot has lower efficiency.}
    \label{single_efficiency}
\end{Figure}

\begin{figure*}[!htbp]
    \centering
    \includegraphics[width=\textwidth]{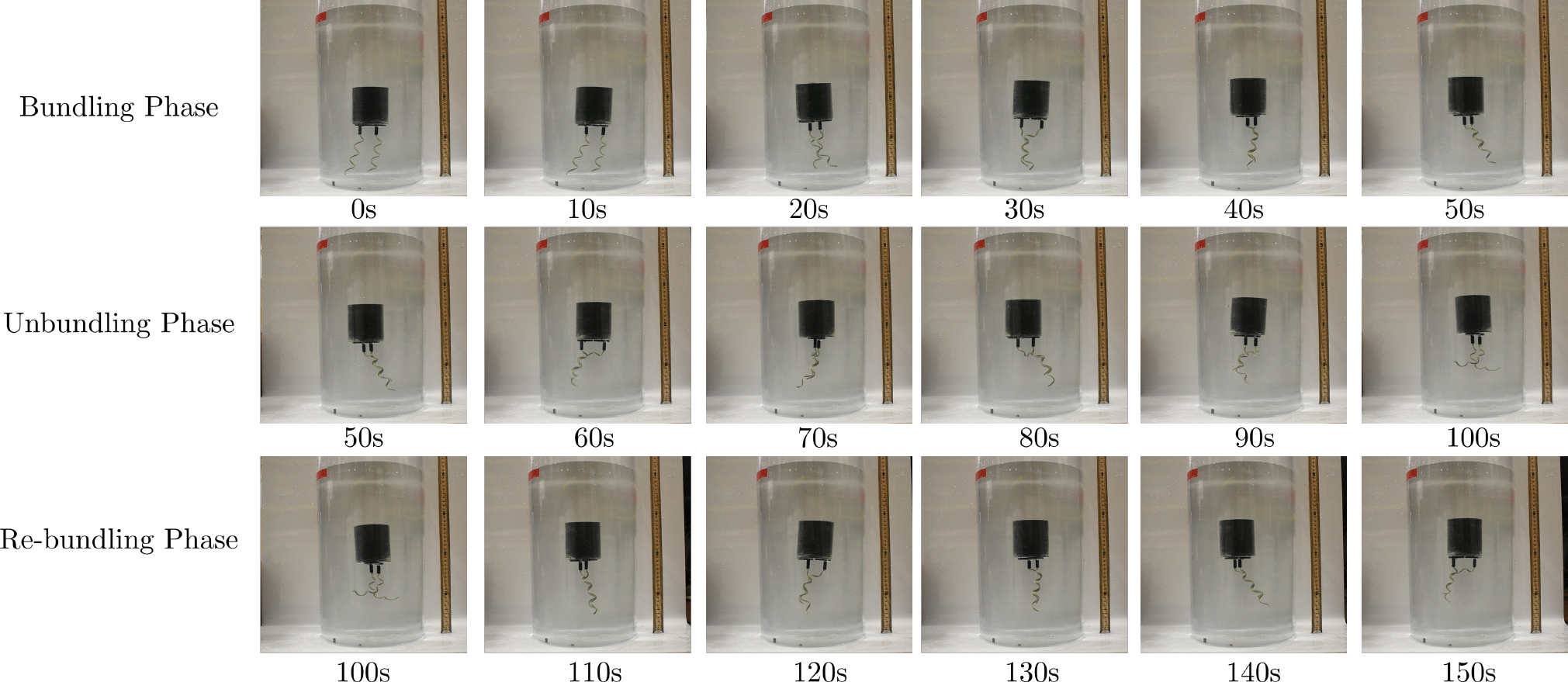}
    \captionof{figure}{Periodic bundling, unbundling, and re-bundling phase. The snapshot is for the case where $\bar\lambda = 5$ at $50$ rpm. Without any change in the rotational velocity input, and due to the contact-initiated friction and slippage, the unbundling occurs. Leading to cyclical behavior of bundle and unbundle of flagella, which was not captured in the simulation.}
    \label{fig:BundleUnbundle}
\end{figure*}

Next, we employ the simulation tool to comparatively explore the propulsive forces of the two types of systems -- single-flagellated and multi-flagellated. Fig.~\ref{singlecompare}.b shows the nondimensionalized propulsive force as a function of total angular velocity at three different pitch values for both the single-flagellated and multi-flagellated cases. Propulsive force is defined as the $x$-component (direction of motion of the robot) of the force exerted on the head (\textbf{Eq. 15, Supplementary methods section S3}). The propulsive force for the multi-flagellated robot was divided by the number of flagella. The propulsive force generated by the robot is small (on the order of $10^{-3}$ N), which makes it difficult to measure experimentally. Our robotic platform does not have a force sensor; therefore, we use simulations to analyze the propulsive force and efficiency of the soft robot. The trend is qualitatively different between the two cases. For single-flagellated robot buckles at a critical angular velocity (indicated by star symbols), and its propulsive force dramatically drops at that point. 
The multi-flagellated robot does not exhibit buckling behavior even at larger angular velocities; the flagella bundle instead. Even though the computational simulation cannot accurately account for the physical friction, Fig.~\ref{singlecompare}.b shows that buckling for multi-flagellated mechanism is not purely hydrodynamic interactions unlike the single flagellated mechanism~\cite{kim2004hydrodynamic}. 
A point of note is the relatively larger propulsive force per flagellum in the single-flagellated robot prior to buckling than in the multi-flagellated robot. The propulsive force depends on the deformed shape of the flagella, and this shape differs between the single-flagellated case (no bundling, only buckling) and the multi-flagellated case (prominent bundling). However, the propulsive force is larger in the multi-flagellated system beyond the critical threshold for buckling in a single-flagellated robot. In short, a single-flagellated robot generates a larger propulsive force per flagellum; however, its propulsion is limited by instability. The non-monotonic dependence of propulsion force on the geometry of the flagella (pitch) is observed in the multi-flagellated case. The robot with $\bar \lambda=5$ generates the largest force and $\bar \lambda=7$ generates the least; $\bar \lambda=9$ falls in between. In contrast, a single-flagellated robot generates more propulsion as the pitch decreases. However, this observation is true only for the range of parameters explored in this study. Prior works~\cite{Rodenborn2013} show non-monotonic dependence of propulsion on the pitch of the flagellum; however, the flagellum was assumed to be rigid.

The observations on propulsive force lead us to address the efficiency of the flagellated robots using numerical simulations. Fig.~\ref{single_efficiency} shows the variation of efficiency with the total angular velocity at three pitch values in both cases. Efficiency is defined as $\eta = \frac{\boldsymbol{f}^\mathrm{head}_h\cdot r_h}{\boldsymbol{T}_h}$, where $\boldsymbol{f}^\mathrm{head}_h$ is the hydrodynamic force on the head, and ${\boldsymbol{T}_h}$ is the torque due to rotation of the head. Qualitatively, efficiency is a measure of the ratio of the propulsive force and the torque exerted by the motor. At smaller values of angular velocity,  multi-flagellated robot decreases in efficiency. This decrease is particularly prominent at small pitch values and, thus, high interaction between flagella. Efficiency is the highest when $\bar \lambda=7$ and lowest at $\bar \lambda = 5$, which further signifies the nonlinear nature of the problem. If the robot has a single flagellum, there is no bundling, and the shape of the flagellum remains almost helical until the threshold for buckling. Therefore, the efficiency remains almost constant as a function of angular velocity before buckling. The efficiency drops to almost zero post-buckling.

\subsection{Bundling and unbundling sequence}
\label{sec:bundling}

In this section, we report the observed sequence of bundling and unbundling found in the experiment. As stated in Section~\ref{sec:validation}, bundling was most exhibited at : 1) higher rotational velocity, 2) lower pitch value. While bundling behavior is captured in both simulation and experiment, unbundling was only captured in the experiment. Previous research on unbundling that does not involve a directional change in the rotation of one or more flagella is limited. Reigh et al. found a stable bundled region characterized by flagella anchor distance and applied torque differences for a multi-flagellated system. The authors of the paper show simulation results on how slippage affects the unbundling process and suggests that slippage may lead to tumbling behavior of bacterial locomotion~\cite{Reigh2013-vl}. 

In Fig.~\ref{fig:BundleUnbundle}, the robot at $50$ rpm with $\bar\lambda = 5$ exhibits bundle and unbundle phase. This cyclical behavior was shown throughout at $40$ - $70$ rpm for the case with $\bar\lambda = 5$. As expected, the initial bundling, while going through the transient phase, occurs for an extended period  ($50$ sec) than the re-bundling phase to get back to a fully bundled state ($10$ sec). After both flagella fully bundle, the unbundling starts. The unbundling process is shown as one of the flagella starting to slip along the bundled helix. After approximately $50$ sec, the bundle reaches a fully unbundled shape. Snapshot at $50$ s - $90$ s and $110$ s - $150$ s on Fig.~\ref{fig:BundleUnbundle} displays that the unbundling phase repeats in the same geometric manner after the full bundle occurs. We suspect the reason for this to be the friction between the two flagella and a possible mismatch in torque due to experimental limitations. After the bundle has been formed, the bundle starts to skew as the contact friction increases. One flagellum tries to get out of the bundle while the friction tries to hold the bundle. As one flagellum slowly escapes from the bundle, the part with friction holds the flagellum that tries to escape from the bundle. However, the unbundled part starts pumping the remaining flagellum out by repeated buckling of the escaping flagellum's unbundled region. 

\section{Conclusion}
In conclusion, we presented a multi-flagellated soft robotic platform and a numerical simulation method. These tools were used to explore the relationship between the motion of the robot and the geometry of the flagella. Both the experiments and simulations could capture the bundling behavior of the elastic flagella caused by long-range hydrodynamic interaction. Bundling is only possible if the flagella are flexible and there is long-range interaction by flows induced by distant parts of the flagella. Prior works often neglected the flexibility of the flagella or ignored the long-range hydrodynamics in favor of a simplified (but computationally cheap) resistive force theory-based hydrodynamic model~\cite{Gray1955}. Our study emphasizes the need to accurately capture these two ingredients -- flexibility and long-range hydrodynamics -- in modeling bacteria and robots inspired by them. The simulation tool can successfully capture both elements.

The accuracy of the simulation, when compared to the experiment, was reasonable, where our metrics for comparison were the translational velocity of the robot and the angular velocity of the head. The lack of an accurate friction model was identified as the main reason behind any mismatch between experiments and simulations and a potential contributing factor to unbundling, which simulation could not capture. This simulation and the experimental platform provide a basis for the future development of multiple flagella-based robots. The results were presented in a nondimensional format. As long as the dimensionless groups are the same (e.g., Reynolds number is low), they apply to robots and organisms of any size.

The robotic platform can be easily re-purposed to explore tumbling -- change in the direction of swimming -- when one flagellum rotates in a direction opposite to the other flagellum. In addition, the robot can be improved by integrating sensors and an inertial measurement unit (IMU) to achieve 3D trajectory control of the robot.

Directions for future work include (1) analysis of the tumbling and unbundling behavior, (2) incorporation of a physically accurate friction model, (3) investigation of the role of head flexibility and geometry, (4) formulation of control policy along with hardware improvement for an autonomous robot. Despite these limitations, we are close to realizing palm-sized flagellated robots that are simple in design (few moving parts) and control (angular velocity is the only control input) yet functional (i.e., capable of following 3D trajectory by bundling and tumbling). With the ongoing advancements in biological discoveries, mechanical experiments, theories, and simulation, we envision that our research can lead to a small, simple, cheap, but functional robot that could fully capture the locomotion of bacteria. 

\section*{Acknowledgements} 
We are grateful for the financial support from the National Science Foundation (Award numbers: IIS-1925360, CMMI-2053971, CMMI-2101751, CAREER-2047663) and the Henry Samueli School of Engineering and Applied Science, University of California, Los Angeles.

\section*{Conflicts of interest}
The authors declare that they have no conflict of interest.


\end{multicols}



\section*{Supplementary material}

## Supplementary Methods

Full details of physics-based simulation for multi-flagellated soft robot.

**S1. The geometry choice of artificial flagella and robot body.** Bacterial flagella are known to exhibit polymorphic transformation. Polymorphic transformation can be induced through environmental changes in pH and ionic strength, temperature, reaction towards organic solvents, reaction towards the electric field or mechanical force (1). However, our paper simplifies the problem by investigating pre-formed elastic helical flagella. Current understanding of the polymorphic transformation of bacterial flagella is limited as well as the material that would make it possible requires further investigation to enable more physically appropriate modeling of bacterial behavior. However, our simplification is still within the range of bacterial flagella. Figure S1 displays different flagella geometry of different bacteria strains and polymorphic states in a blue triangle shape with the name and our experimental geometry in a red circle. Our experiment lies reasonably within the geometric range of bacterial flagella in nature. The examples of the species were adapted from Rodenborn et al. (2) Table 1.

Another design concern for the robot was the shape of the head. The actual body of the bacteria is a very flexible ellipsoidal shape. However, the modeling of structure and experimental realization for the cell body of flagellated bacteria are limited (3, 4). Our design is relevantly chosen to ensure concentric rotation of the flagella for each motor with minimal leaking from the gap. The concentric rotation of flagella with minimal gap would be complex with the flexible ellipsoidal body. Furthermore, utilizing our current design, even though it may not be soft and ellipsoidal, we were able to capture the counter-rotation of the robot head and the bundling of the flagella. In future works, we can improve our model and the modeling methods to gain an advantage using the shape of the cell body as well as the flagella interaction (5).

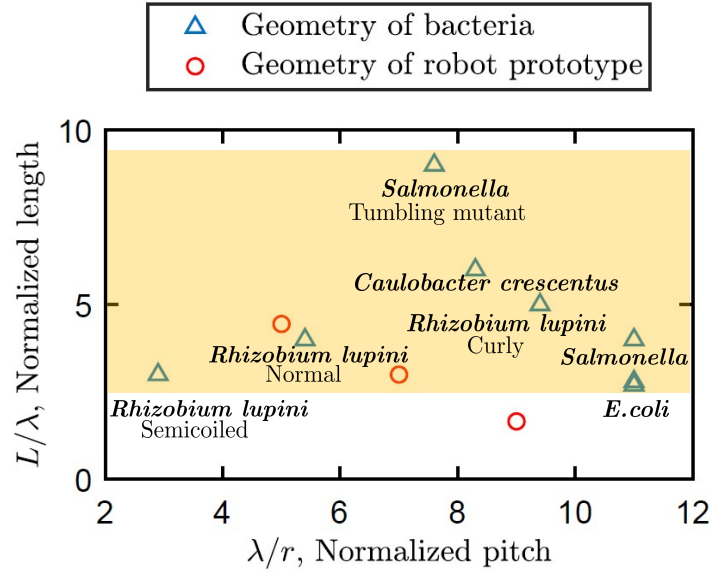

**Fig. S1.** Comparison of existing bacterial flagella geometry and the artificial flagella geometry of the robot prototype. The shaded area shows the range of the bacterial flagella geometry.

**S2. Elastic Strains, Forces and Jacobians.** In order to understand the Discrete Elastic Rod algorithm (DER), it is essential to understand the elastic strains, associated energy, gradient and hessian of the energy term. As described in the main text, there are three major elastic strains that a Kirchoff rod-based formulation take into account : bending, twisting, and stretching. Bending and twisting strains are node-based quantities while stretching strain is an edge-based quantity. Bending is computed using the curvature binormal vector at each node:

$$(\kappa \mathbf{b})_k = \frac{2 \boldsymbol{t}^{k-1} \times \boldsymbol{t}^k}{1 + \boldsymbol{t}^{k-1} \cdot \boldsymbol{t}^k} \qquad [1]$$

The magnitude of this vector is $2\tan(\phi_k/2)$ where $\phi_k$ is the turning angle shown in Figure 4. The curvature vector (i.e. bending strain) at the $k$-th node is then

$$\boldsymbol{\kappa}_k = \left((\kappa \mathbf{b})_k \cdot \mathbf{d}_2^k, -(\kappa \mathbf{b})_k \cdot \mathbf{d}_1^k\right). \qquad [2]$$

The twist at each node is

$$\tau_k = \theta^k - \theta^{k-1} + m_{k,\text{ref}}, \qquad [3]$$

where $m_{k,\text{ref}}$ represents the reference twist (i.e. twist associated with the reference frame) and can be calculated from the reference frames (6). In order to account for the rotation of the motor, we included a "natural" twist angle, $\tau_{\text{motor}} = \omega_T \cdot t$ (where $t$ represents time), to the expression for twist at the two nodes representing the motors ($\mathbf{x}_{\text{m1}}$, and $\mathbf{x}_{\text{m2}}$) shown in Figure 4. Then the new equation for integrated twist at $\mathbf{x}_{\text{m1}}$ and $\mathbf{x}_{\text{m2}}$ becomes

$$\tau_k = \theta^k - \theta^{k-1} + m_{k,\text{ref}} - \tau_{\text{motor}}, \quad [4]$$

where $\tau_{\text{motor}}$ is zero everywhere except at m1 and m2. Axial stretching ($\epsilon^k$) is an edge-based quantity, which can be represented as

$$\epsilon^k = \frac{|\mathbf{x}_{k+1} - \mathbf{x}_k|}{|\overline{\mathbf{e}^k}|} - 1, \quad [5]$$

where $|\overline{\mathbf{e}^k}|$ denotes undeformed maginitude of the $k$-th edge. The energy term associated with the elastic strains can be calculated to be

$$E_k^s = \frac{1}{2} EA (\epsilon^k)^2 |\overline{e^k}|, \quad [6]$$

$$E_k^b = \frac{1}{2} EI (|\boldsymbol{\kappa_k} - \boldsymbol{\kappa_k^0}|)^2 \frac{1}{\bar{l}_k}, \quad [7]$$

$$E_k^t = \frac{1}{2} GJ \tau_k^2 \frac{1}{\bar{l}_k}, \quad [8]$$

where $EA = E\pi r_0^2$ is the stretching stiffness, $EI = E\pi r_0^4/4$ is the bending stiffness, $GJ = G\pi r_0^4/2$ is the twisting stiffness, $G = E/(2(1+\nu))$ is the shearing modulus, and $\bar{l}_k$ is the reference Voronoi length: $\bar{l}_k = \frac{1}{2}\left(|\overline{e^{k-1}}| + |\overline{e^k}|\right)$. The gradient of these energy terms with respect to the degree of freedom provides us with the forces associated with the energy that is required for obtaining the equation of motion. To make sure that our iteration is implicit, we can define the Jacobian terms for the components of equations of motion as,

$$\mathbb{J}_{km}^{\text{inertia}} = \frac{m_k}{\Delta t^2} \delta_{km}, \quad [9]$$

$$\mathbb{J}_{km}^{\text{elastic}} = \frac{\partial^2 E_{\text{elastic}}}{\partial \mathbf{q}_k \partial \mathbf{q}_m}, \quad [10]$$

$$\mathbb{J}_{km}^{\text{ext}} = -\frac{\partial f_k^{\text{ext}}}{\partial \mathbf{q}_m}, \quad [11]$$

where $\delta_{km}$ is the Kronecker delta function ($\delta_{km} = 1$ if $k = m$; otherwise $\delta_{km} = 0$). The expressions for the Jacobian terms associated with the elastic forces are available in the literature (6).

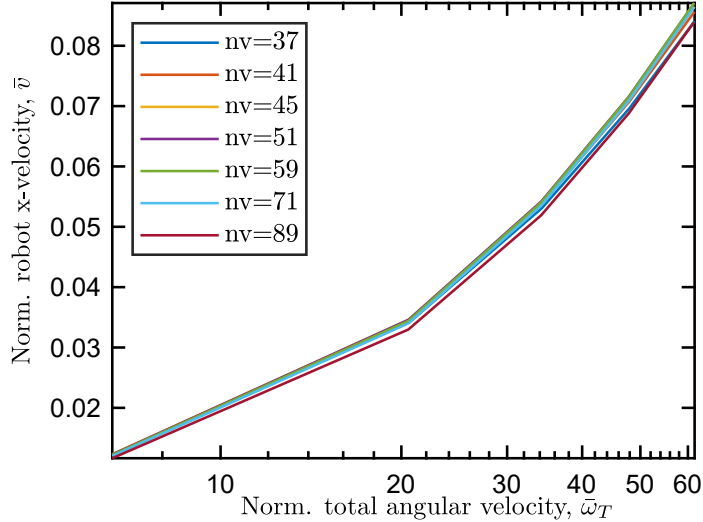

**Fig. S2.** Convergence study done for $\bar{\lambda} = 7$. Nodes numbers of 7 different cases were tested to find an optimal discretization.

In order to decide the level of discretization we have conducted a convergence study for $\bar{\lambda} = 7$ using normalized robot velocity compared to normalized total angular velocity as shown in Figure S2. The simulation data were collected for a total of 1000 seconds of simulation time with 5 rpm increments from 5rpm to 50rpm. As shown in the figure, the velocity converges very fast even with the number of nodes as few as 37. In order not to sacrifice the speed and accuracy of the coupled simulation, we used 59 nodes, ($|e| = 5e^{-3}$). Corresponding number of nodes for other geometric cases are $\bar{\lambda} = 9 : \mathbf{5}5$ nodes, $\bar{\lambda} = 5 : 69$ nodes.

**S3. Hydrodynamics model.** We used the Regularized Stokeslet Segments method for the hydrodynamic force on the flagella and Stokes law for the hydrodynamic force on the robot head. Built on the method of regularized Stokeslets, RSS method is beneficial to reduce the sensitivity of the velocity field to the regularization parameter due to its numerical treatment of a weakly singular integral. Cortez presented this method with the assumption that the force field along a filament is piece-wise linear and suggested the possibility of application to piece-wise quadratic or higher degree polynomial (7). Importantly, RSS method accounts for the long range hydrodynamic interaction among flows induced by different nodes on the flagellum. This interaction is ignored by widely used simplified methods, also known as Resistive Force Theory (8).

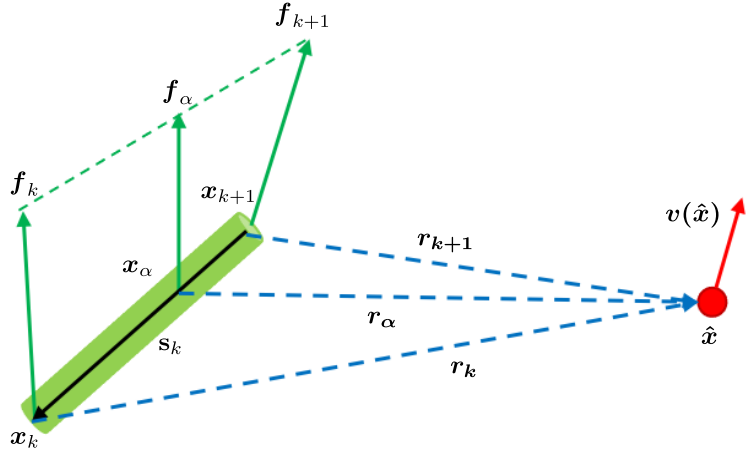

**Fig. S3.** Notations and discrete schematic of the flow at $\hat{x}$ generated by a line segment.

Referring to Figure S3, RSS provides a relationship between the velocity at a point ($v(\hat{x})$ in Figure S3) and the forces applied by each node on the fluid such that

$$8\pi\mu v(\hat{x}) = \sum_{k=0}^{N-2}(M_1^k f_k^h + M_2^k f_{k+1}^h). \quad [12]$$

where $f_k^h$ is the force vector of size 3 that represents the force applied by the $k$-th node onto the fluid. This is equal and opposite to the hydrodynamic force onto the $k$-th node. The hydrodynamic force on the $k$-th node is the $4k, (4k+1)$, and $(4k+2)$-th elements of the hydrodynamic force vector $f^h$ of size $(4N-1)$ in equation 3 in the main text. The matrices (size $3 \times 3$) $M_1^k$ and $M_2^k$ are

$$M_2^k = |\mathbf{s}_k|((T_{1,-1}^{k,k+1} + \epsilon^2 T_{1,-3}^{k,k+1})\mathbb{I} + T_{1,-3}^{k,k+1}(\mathbf{r}_k \mathbf{r}_k^T) + \\ T_{2,-3}^{k,k+1}(\mathbf{r}_k \mathbf{s}_k^T + \mathbf{s}_k \mathbf{r}_k^T) + T_{3,-3}^{k,k+1}(\mathbf{s}_k \mathbf{s}_k^T)), \quad [13a]$$

$$M_1^k = |\mathbf{s}_k|((T_{0,-1}^{k,k+1} + \epsilon^2 T_{0,-3}^{k,k+1})\mathbb{I} + T_{0,-3}^{k,k+1}(\mathbf{r}_k \mathbf{r}_k^T) + \\ T_{1,-3}^{k,k+1}(\mathbf{r}_k \mathbf{s}_k^T + \mathbf{s}_k \mathbf{r}_k^T) + T_{2,-3}^{k,k+1}(\mathbf{s}_k \mathbf{s}_k^T)) - M_2^k, \quad [13b]$$

where, as shown in Figure S3, $\hat{x}$ is the point of evaluation, $\epsilon$ is regularization parameter, $\mathbf{r}_k = \hat{\mathbf{x}} - \mathbf{x}_k$, $\mathbf{s}_k = \mathbf{x}_k - \mathbf{x}_{k+1}$, $\mathbb{I}$ is 3-by-3 identity matrix, and the scalar quantities denoted by $T$ (e.g. $T_{1,-1}^{k,k+1}$) are described next.

$$T_{0,-1}^{k,k+1} = \frac{1}{|\mathbf{s}_k|} \log[|\mathbf{s}_k|R + (\mathbf{x}_\alpha \cdot \mathbf{s}_k)]\Big|_0^1, \quad [14a]$$

$$T_{0,-3}^{k,k+1} = -\frac{1}{R[|\mathbf{s}_k|R + (\mathbf{x}_\alpha \cdot \mathbf{s}_k)]}\Big|_0^1, \quad [14b]$$

$$T_{1,-1}^{k,k+1} = \frac{R}{(|\mathbf{s}_k|)^2}\Big|_0^1 - \frac{(\mathbf{x}_0 \cdot \mathbf{s}_k)}{(|\mathbf{s}_k|)^2} T_{0,-1}^{k,k+1}, \quad [14c]$$

$$T_{1,-3}^{k,k+1} = -\frac{1}{R(|\mathbf{s}_k|)^2}\Big|_0^1 - \frac{(\mathbf{x}_0 \cdot \mathbf{s}_k)}{(|\mathbf{s}_k|)^2} T_{0,-3}^{k,k+1}, \quad [14d]$$

$$T_{2,-3}^{k,k+1} = -\frac{\alpha}{R(|\mathbf{s}_k|)^2}\Big|_0^1 + \frac{1}{(|\mathbf{s}_k|)^2} T_{0,-1}^{k,k+1} - \\ \frac{(\mathbf{x}_0 \cdot \mathbf{s}_k)}{(|\mathbf{s}_k|)^2} T_{1,-3}^{k,k+1}, \quad [14e]$$

$$T_{3,-3}^{k,k+1} = -\frac{\alpha^2}{R(|\mathbf{s}_k|)^2}\bigg|_0^1 + \frac{2}{(|\mathbf{s}_k|)^2}T_{1,-1}^{k,k+1} - \frac{(\mathbf{x}_0 \cdot \mathbf{s}_k)}{(|\mathbf{s}_k|)^2}T_{2,-3}^{k,k+1}$$  [14f]

where $\mathbf{x}_\alpha = \mathbf{x_k} - \alpha \mathbf{s_k}$, and $R = \sqrt{|\mathbf{x}_\alpha|^2 + \epsilon^2}$. Equation 12 can be used to formulate $3N$ equations (3 per node) that relate the velocities at each node with the forces applied by all the other nodes. Knowing the velocity of each node at the beginning of the time step in DER, this linear system of equations can be solved to obtain the forces and compute the hydrodynamic force vector $\mathbf{f}^h$. Since the gradient of the forces with respect to the DOF vector is not available, this force is treated explicitly (Euler forward) in the simulation scheme. Complete details are found in Refs. (7) and (9).

Previously, MRS method prevailed for the analysis in low Reynolds hydrodynamics using Stokeslet methods, however, due to its dependence on the distance between contiguous cutoff functions, the choice of regularization parameter $\epsilon$ limited the accuracy of calculation. However, RSS method accounts for a continuum of regularized forces therefore decouples the necessity between discretization and the regularization. The regularization parameter used in RSS method $\epsilon$ for flagella could be interpreted as the radius of the slender filaments. Based on the analysis shown from Cortez (7) we used the regularization value of $\epsilon = 1.031 \cdot r_0 = 0.00165$ (m).

We now turn to the computation of the hydrodynamic forces on the head ($\mathbf{f}^{\text{head}}$ in equation 3 in the main text). The middle node along the entire structure ($\mathbf{x}_h$ in Figure 4) represents the head. As the head is translating (quantified by the velocity of $\mathbf{x}_h$), the viscous medium exerts a drag force onto it. The head is also rotating (quantified by the angular velocity of the head, $\boldsymbol{\omega}_h$) and the viscous fluid applies a torque to resist that rotation. We applied Stokes' law to model the hydrodynamic drag. Since Stokes' law is meant for spherical bodies and the robotic head is cylindrical, we used two numerical prefactors as fitting parameters as discussed next.

The hydrodynamic force on the head ($\mathbf{x}_h$) at time $t = t_{i+1}$ is

$$\mathbf{f}_h^{\text{head}} = -C_t \cdot 6\pi\mu r_h \left[\frac{\mathbf{x}_h(t_{i+1}) - \mathbf{x}_h(t_i)}{\triangle t}\right],$$  [15]

where $C_t$ is a numerical prefactor to account for the non-spherical shape of the head, $r_h$ is the radius of the head (see Table 1), and $\mathbf{x}_h$ is the position of the head node that can be extracted from the DOF vector, $\mathbf{q}$. If the head is the $h$-th node in the structure, the vector $\mathbf{f}_h^{\text{head}}$ of size 3 represents $(4h-3), (4h-2)$, and $(4h-1)$-th elements of the vector $\mathbf{f}^{\text{head}}$ in Eq.3.

The torque due to rotation of the head is $\mathbf{T}_h = -C_r \cdot 8\pi\mu r_h^3 \boldsymbol{\omega}_h$, where $C_r$ is a numerical prefactor due to the non-spherical shape of the head and $\boldsymbol{\omega}_h$ is the angular velocity of the head. The angular velocity at $t = t_{i+1}$ can be computed from the velocities of the two neighboring nodes $\mathbf{x}_{h-1}$, and $\mathbf{x}_{h+1}$ (see Figure 4) such that

$$\omega_h = \frac{1}{|\mathbf{x}_{h+1}(t_{i+1}) - \mathbf{x}_{h-1}(t_{i+1})|^2}$$
$$|(\mathbf{x}_{h+1}(t_{i+1}) - \mathbf{x}_{h-1}(t_{i+1})) \times [\dot{\mathbf{x}}_{h+1} - \dot{\mathbf{x}}_{h-1}]|,$$  [16]

where $\dot{\mathbf{x}}_{h+1} = \frac{\mathbf{x}_{h+1}(t_{i+1}) - \mathbf{x}_{h+1}(t_i)}{\triangle t}$ is the velocity of the $(h+1)$-th node, $\dot{\mathbf{x}}_{h-1} = \frac{\mathbf{x}_{h-1}(t_{i+1}) - \mathbf{x}_{h-1}(t_i)}{\triangle t}$ is the velocity of the $(h-1)$-th node, and $\times$ represents vector cross product. The hydrodynamic torque is implemented in the simulation as a force-couple, i.e. a force on the node $\mathbf{x}_{h-1}$ and an equal but opposite force on the node $\mathbf{x}_{h+1}$. It turns out that, in case of the specific problem studied in this paper, the magnitude of each force in the force-couple can be approximated to a very good degree as $\mathbf{T}_h/|\mathbf{x}_{h+1}(t_{i+1}) - \mathbf{x}_{h-1}(t_{i+1})|$ and the angle between two vectors in the right-side of equation 16 is 90°. The direction of the force can be approximated to be equal to $[\dot{\mathbf{x}}_{h+1} - \dot{\mathbf{x}}_{h-1}]$. The reason behind these approximations is they result in a simplified expression for the forces and allow us to take the gradient with respect to the DOFs (so that the forces are incorporated into the simulation implicitly). The resulting force on the $(h+1)$-th node is

$$\mathbf{f}_{h+1}^{\text{head}} = -C_r \cdot 8\pi\mu \frac{r_h^3}{|x_{h+1}(t_{i+1}) - x_{h-1}(t_{i+1})|^2}[\dot{\mathbf{x}}_{h+1} - \dot{\mathbf{x}}_{h-1}].$$  [17]

The force on $\mathbf{x}_{h-1}$ is

$$\mathbf{f}_{h-1}^{\text{head}} = -\mathbf{f}_{h+1}^{\text{head}}.$$  [18]

Equations 15, 17, and 18 are used to calculate the hydrodynamic forces on the head and populate the $(4N-1)$-sized $\mathbf{f}^{\text{head}}$ vector. Note that this vector has only 9 non-zero elements (resulting from the forces on 3 nodes).

We fitted the parameters $C_t$ and $C_r$ with the experiment velocity along the x-axis which is defined in Figure 5 and head rotation speed for $\bar{\lambda} = 7$ case. The values for $C_t$ varied from $4.9 \sim 5.1$ and $C_r$ varied from $0.6 \sim 1.5$ for the simulation sample data. The mean total least squared error from the experimental values were the smallest for the head angular velocity when $C_r = 0.9$, and $C_t = 4.1$ for the x-velocity. We used these values for the head hydrodynamics for analysis of other geometric cases for flagella as well ($\bar{\lambda} = 5, \bar{\lambda} = 9$).

**S4. Contact model.** In this section, the constraint-based contact forces are explained based on the non-penetrative condition between the two edges; we refer the reader to Ref. (10) for complete details. Figure S4 shows two edge segments undergoing collision. We denote the edge segments as $S_k = (\mathbf{x}_k, \mathbf{x}_{k+1})$ and $S_m = (\mathbf{x_m}, \mathbf{x}_{m+1})$, where $\mathbf{x_k}$, $\mathbf{x}_{k+1}$, $\mathbf{x}_m$, and $\mathbf{x}_{m+1}$ can be extracted from our DOF vector $\mathbf{q}$. In order to formulate a non-penetrative condition, penetration depth ($\epsilon_{k,m}$) is defined as the difference between the minimum Euclidean distance $\delta_{k,m}^{min}$ between $S_m$ and $S_k$ and sum of the radii of the segment $S_k$ and $S_m$:

$$\epsilon_{k,m} = 2r_0 - \delta_{k,m}^{min}. \qquad [19]$$

In our case, the radius of all the segments are the same and therefore the sum of the radii is always $2r_0$.

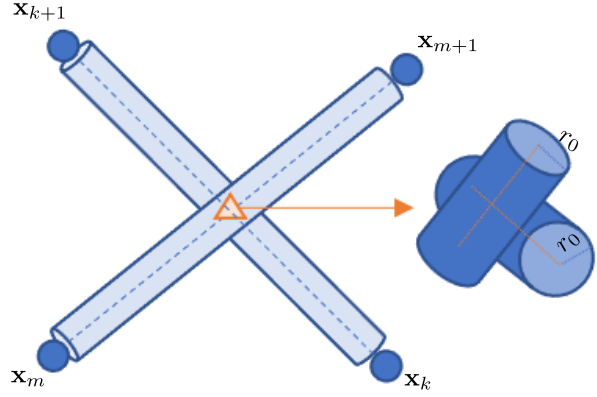

**Fig. S4.** Schematic of contact between two line segments $S_k$ and $S_m$; orange triangle represents point of contact, $r_0$ represents the radius of each rod. The minimum distance two segments cannot be smaller than $2r_0$.

A contact is detected when $\epsilon_{k,m} > 0$. The schematic shown in Figure S4 visualizes the contact condition. After detecting contact, iterations to find interference-free configuration needs to be executed. Using the detected points, the collision displacements are weighted with the barycentric coordinates of the collision to account for the conservation of mass such that

$$\begin{aligned}\Delta\mathbf{x}_k &= -\frac{1}{2}\mathbf{n}_{km}w_k \\ \Delta\mathbf{x}_{k+1} &= -\frac{1}{2}\mathbf{n}_{km}(1-w_k) \\ \Delta\mathbf{x}_m &= \frac{1}{2}\mathbf{n}_{km}w_m \\ \Delta\mathbf{x}_{m+1} &= \frac{1}{2}\mathbf{n}_{km}(1-w_m).\end{aligned} \qquad [20]$$

Here, the value of $\frac{1}{2}$ represents the barycentrical ratio of masses with all points having same masses, $\mathbf{n}_{km}$ represents the minimum distance vector between $S_k$ and $S_m$, $w_k$ represents the barycentric coordinate of the contact point on the segment. Iterative process over all the detected contact positions and summing the displacement points for each $\mathbf{x}_k$, the displacements with mass conservation could be obtained. We use the penetration depth and compare it with the error tolerance until $\epsilon_{k,m}^{min}$ < tolerance for all the detected contact points. The contact force applied at $k$-th node due to collision between the contact segments $S_k$ and $S_m$ can be evaluated using the $\Delta\mathbf{x}_k$ values and is represented as

$$\mathbf{f}_k^c = \frac{1}{\Delta t^2}\Delta\mathbf{x}_k m_k, \qquad [21]$$

where $\Delta\mathbf{x}_k$ represents weighted collision displacements with mass conservation consideration, $\Delta t$ represents the time discretization, $m_k$ represents the mass of the point (10).

While formulating the DER with contact, we found out that the level of discretization is limited by contact function. Contact method by Spillman and Teschner (10) considers the diameter of the rod so that when two nodes are within the boundary of its diameter, contact could be detected. Therefore, contact limits our length of discretized segment $|e|$ to be always greater than the diameter of the rod $2r_0$.

While the limitation in level of discretization could be a limitation, we used RSS method with our particular choice of the regularization parameter that decouples the viscous force along a line segment from discretization to overcome the limitation. It is shown by Cortez that the level of discretization has insignificant effect on the swimming speed and the trajectory waveform (7).

**Movie S1. Comparison video of multi-flagellated robotic platform and simulation with flagella geometry variation**



\end{document}